\documentclass[10pt,twocolumn,letterpaper]{article}

\usepackage{cvpr}
\usepackage{times}
\usepackage{epsfig}
\usepackage{graphicx}
\usepackage{amsmath}
\usepackage{amssymb}
\usepackage{float}
\usepackage{multirow}

\newcommand{\norm}[1]{\left\lVert#1\right\rVert}
\DeclareMathOperator*{\argmin}{argmin}

\cvprfinalcopy 

\def\cvprPaperID{3796} 

\ifcvprfinal\fi
\begin{document}

\title{Cascaded Projection: End-to-End Network Compression and Acceleration}

\author{Breton Minnehan\\
Rochester Institute of Technology\\
{\tt\small blm2144@rit.edu}
\and
Andreas Savakis\\
Rochester Institute of Technology\\
{\tt\small andreas.savakis@rit.edu}
}

\maketitle

\begin{abstract}
We propose a data-driven approach for deep convolutional neural network compression that achieves high accuracy with high throughput and low memory requirements. 
Current network compression methods either find a low-rank factorization of the features that requires more memory, or select only a subset of features by pruning entire filter channels. 
We propose the Cascaded Projection (CaP) compression method that projects the output and input filter channels of successive layers to a unified low dimensional space based on a low-rank projection. 
We optimize the projection to minimize classification loss and the difference between the next layer's features in the compressed and uncompressed networks. 
To solve this non-convex optimization problem we propose a new optimization method of a proxy matrix using backpropagation and Stochastic Gradient Descent (SGD) with geometric constraints.  
Our cascaded projection approach leads to improvements in all critical areas of network compression: high accuracy, low memory consumption, low parameter count and high processing speed.
The proposed CaP method demonstrates state-of-the-art results compressing VGG16 and ResNet networks with over 4$\times$ reduction in the number of computations and excellent performance in top-5 accuracy on the ImageNet dataset before and after fine-tuning. 
\end{abstract}

\section{Introduction}
\label{sec:intro}
\begin{figure}[t]
   \includegraphics[width=2.75in]{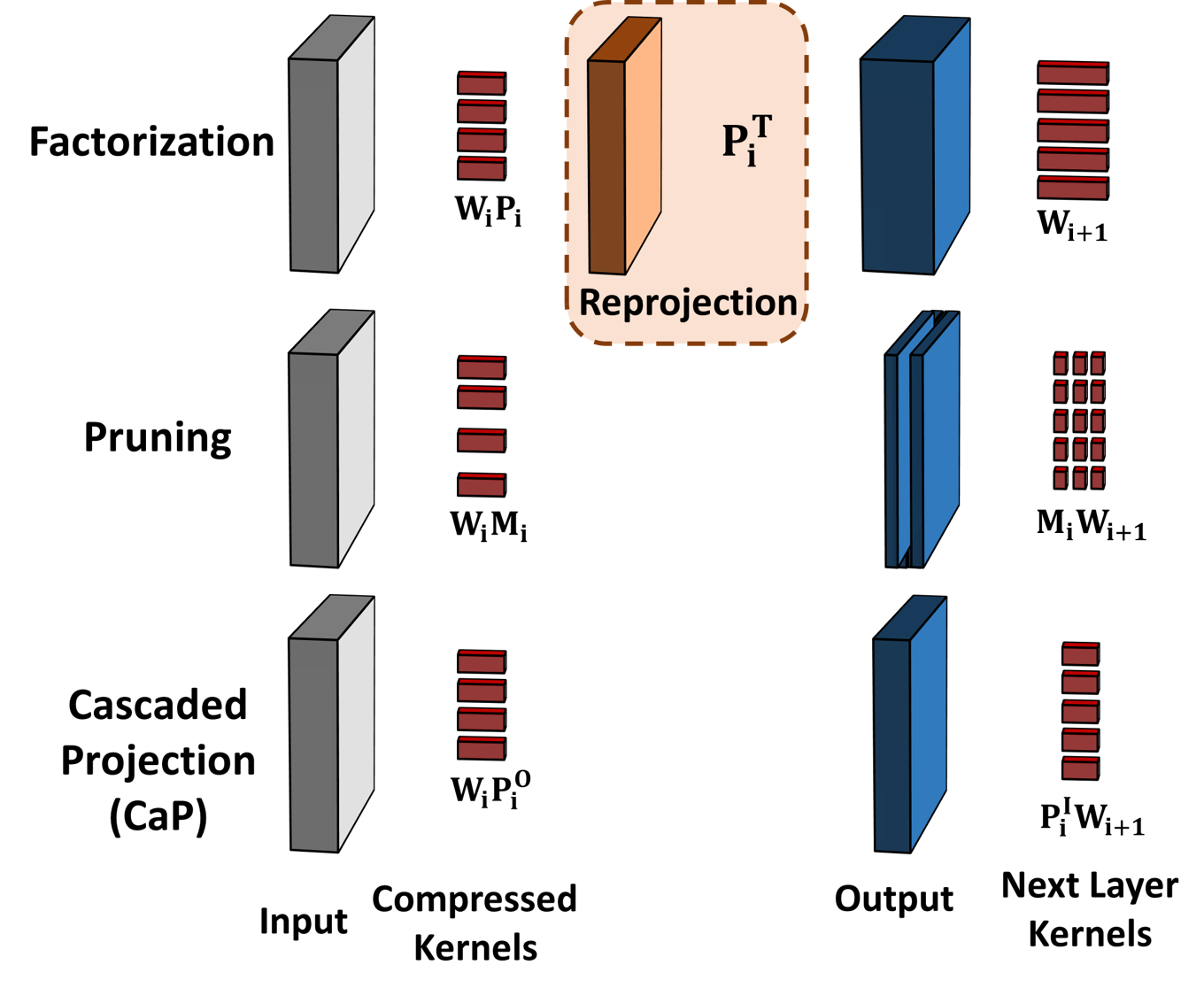}
   \caption{Visual representation of network compression methods on a single CNN layer. Top row: Factorization compression with a reprojection step that increases memory. Middle row: Pruning compression where individual filters are removed. Bottom row: Proposed CaP method which forms linear combinations of the filters without requiring reprojection.
   }
\label{fig:compmethods}
\end{figure}

The compression of deep neural networks is gaining attention due to the effectiveness of deep networks and their  potential applications on mobile and embedded devices.
The powerful deep networks developed today are often overparameterized \cite{denil2013predicting} and require large amounts of memory and computational resources \cite{canziani2016analysis}.
Thus, efficient network compression, that reduces the number of computations and memory required to process images, enables the broader application of deep neural networks.

Methods for network compression can be categorized into four types, based on quantization, sparsification, factorization and pruning. 
In this work we leverage the advantages of factorization and pruning methods, as they are the most popular.
Quantization methods accelerate deep networks and reduce storage by using mixed precision arithmetic and hashing codes \cite{chen2015compressing, courbariaux2014training, han2015deep}. However most of them require mixed precision arithmetic, which is not always available on standard hardware.
Sparsification methods eliminate individual connections between nodes that have minimal impact on the network, however, they are not well suited for current applications because most neural network libraries are not optimized for sparse convolution operations and fail to achieve significant speedup.

Factorization methods \cite{denton2014exploiting, jaderberg2014speeding,lebedev2014speeding,zhang2015efficient} reduce computations by factorizing the network kernels, often by splitting large kernels into a series of convolutions with smaller filters. 
These methods have the drawback of increasing memory consumption due to the intermediate convolution operations.
Such memory requirements pose a problem for mobile applications, where network acceleration is needed most.
Pruning methods  \cite{han2015deep,He_2017_ICCV,li2016pruning,luo2017thin,molchanov2016pruning, srinivas2015data,Yu_2018_CVPR, NIPS2018_7367} compress layers of a network by removing entire convolutional filters and the corresponding channels in the filters of the next layer. 
They do not require feature map reprojection, however they discard a large amount of information when eliminating entire filter channels. 

In this paper, we propose the Cascaded Projection (CaP) compression method which combines the superior reconstruction ability of factorization methods with the multi-layer cascaded compression of pruning methods.
Instead of selecting a subset of features, as is done in pruning methods, CaP forms linear combinations of the original features that retain more information. 
However, unlike factorization methods, CaP brings the kernels in the next layer to low dimensional feature space and therefore does not require additional memory for reprojection.

Figure \ref{fig:compmethods} provides a visual representation of the differences between the three methods: factorization (top row) reprojects to higher dimensional space and increases memory, pruning (middle row) masks filters and eliminates their channels, and our proposed CaP methods (bottom row) combines filters to a smaller number without reprojecting.
Our results demonstrate that by forming filters based on linear combinations instead of pruning with a mask, more information is kept in the filtering operations and better network classification accuracy is achieved. 
The primary contributions of this paper are the following:
\begin{enumerate}
    \vspace{-0.1in}
\item We propose the CaP compression method that finds a low dimensional projection of the feature kernels and cascades the projection to compress the input channels of the kernels in the next layers.
\vspace{-0.1in} 
\item We introduce proxy matrix projection backpropagation, the first method to optimize the compression projection for each layer using end-to-end training with standard  backpropagation and stochastic gradient descent. 
\vspace{-0.1in}
\item Our optimization method allows us to use a new loss function that combines the reconstruction loss with classification loss to find a better solution.
\vspace{-0.1in}
\item The CaP method is the first to simultaneously optimize the compression projection for all layers of residual networks. 
\vspace{-0.1in}
\item Our results illustrate that CaP compressed networks achieve state-of-the-art accuracy while reducing the network's number of parameters, computational load and memory consumption.
\end{enumerate}

\section{Related Work}
\label{sec:background}
The goal of network compression and acceleration is to reduce the number of parameters and computations performed in deep networks without sacrificing accuracy. 
Early work in network pruning dates back to the 1990's \cite{hassibi1993second}. However, the area did not gain much interest until deep convolutional networks became common \cite{krizhevsky2009learning, ksh12, sz15} and 
the redundancy of network parameters became apparent \cite{denil2013predicting}.
Recent works aim to develop smaller network architectures that require fewer resources \cite{howard2017mobilenets, iandola2016squeezenet, yolov3}. 
%

Quantization techniques \cite{chen2015compressing, courbariaux2014training, han2015deep, jacob2017quantization} use integer or mixed precision arithmetic only available on state-of-the-art GPUs \cite{markidis2018nvidia}.
These methods reduce the computation time and the amount of storage required for the network parameters.
They can be applied in addition to other methods to further accelerate compressed networks, as was done in \cite{kim2015compression}.

Network sparsification \cite{liu2015sparse}, sometimes referred to as unstructured pruning, reduces the number of connections in deep networks by imposing sparsity constraints.
The work in \cite{Huang_2018_CVPR} proposed recasting the sparsified network into separate groups of operations where the filters in each layer are only connected to a subset of the input channels.
However, this method requires training the network from scratch which is not practical or efficient. 
Furthermore,  current hardware is not designed for efficient sparse operations making the process less efficient. 

Filter factorization methods reduce 
computations
at the cost of increased memory load for storing intermediate feature maps.
Initial works focused on factorizing the three-dimensional convolutional kernels into three separable one-dimensional  filters \cite{denton2014exploiting, jaderberg2014speeding}.
In \cite{lebedev2014speeding} CP-decomposition is used to decompose the convolutional layers into five layers with lower complexity.
More recently \cite{zhang2015efficient} performed a  channel decomposition that found a projection of the convolutional filters in each layer such that the asymmetric reprojection error was minimized.

Channel pruning methods \cite{li2016pruning,luo2017thin,molchanov2016pruning, srinivas2015data} remove entire feature kernels for network compression.
In \cite{han2015deep} kernels are pruned based on their magnitudes, under the assumption that kernels with low magnitudes provide little information to the network.
Li \etal \cite{li2016pruning} suggested a similar pruning technique based on kernel statistics.
He \etal \cite{He_2017_ICCV} proposed pruning filters based on minimizing the reconstruction error of each layer.
 Luo \etal \cite{luo2017thin} further extended the concepts in \cite{He_2017_ICCV} to prune filters that have minimal impact on the reconstruction of the next layer.
Yu  \etal \cite{Yu_2018_CVPR} proposed Neuron Importance Score Propagation (NISP)
to calculate the importance of each neuron based on its contribution to the final feature representation and prune feature channels that provide minimal information to the final feature representation. 

Other recent works have focused less on finding the optimal set of features to prune and more on finding the optimal amount of features to remove from each layer of the network.
This is important to study because the amount of pruning performed in each layer is often set arbitrarily or through extensive experimentation. 
In \cite{yamamoto2018pcas, Yu_2018_CVPR} the authors propose automatic pruning architecture methods based on statistical measures. 
In \cite{he2018amc,huang2018learning} methods are proposed which use reinforcement learning to learn an optimal network compression architecture. 
Additional work has been done to reduce the number of parameters in the final layers of deep networks \cite{cheng2015exploration}, however the fully connected layer only contributes a small fraction of the overall computations.

\section{Cascaded Projection Methodology}
\label{sec:method}
In this section we provide an in depth discussion of the CaP compression and acceleration method.
We first introduce projection compression when applied to a single layer, and explain the relationship between CaP and previous filter factorization methods \cite{zhang2015efficient}.
One of the main goals of CaP compression is eliminating the feature reprojection step performed in factorization methods. 
To accomplish this, CaP extends the compression in the present layer to the inputs of the kernels in the next layer by projecting them to the same low dimensional space, as shown in Figure \ref{sec:background}.  
Next we demonstrate that, with a few alterations, the CaP compression method can   
perform simultaneous optimization of the projections for all of the layers in residual networks \cite{he2016deep}. 
Lastly we present the core component of the CaP method, which is our new end-to-end optimization method that optimizes the layer compression projections using standard back-propagation and stochastic gradient descent. 

\subsection{Problem Definition}
\label{sub:probDef}
In a convolutional network, as illustrated in the top row of Fig. \ref{sec:background}, the $i^{th}$ layer takes as input a 4-Tensor ${\bf I_{i}}$ of dimension $(n \! \times \! c_{i} \! \times \! h_{i} \! \times \! w_{i})$, where $n$ is the number of images (mini-batch size) input into the network, $c_{i}$ is the number channels in the input and $w_{i}$ and $h_{i}$ are the height and width of the input. 
The input is convolved with a set of filters ${\bf W_{i}}$ represented as a 4-Tensor with dimensions 
$(c_{i+1} \! \times \! c_{i} \! \times \! k \! \times \! k)$,
where $c_{i+1}$ is the number of kernels and $k$ is the spatial dimensions of the kernels, generally 3 pixels. 
In many networks, there is an additional bias, ${\bf b_{i}}$, of dimension $(c_{i+1}\times 1 \times 1 \times 1)$, that is added to each channel of the output. 
More formally, the convolution operation for layer $i$ of a CNN is given as:
\begin{equation}
{\bf O_{i}} = {\bf I_{i}}*{\bf W_{i}}+{\bf b_{i}}
\end{equation}
\noindent where ($*$) is the convolution operator. 
The input to the next layer is calculated by applying a nonlinearity to the output as ${\bf I_{i+1}}= G({\bf O_{i}})$, where $G(\cdot)$ is often a ReLU \cite{nair2010rectified}.

Network compression aims to reduce the number of filters 
so that the classification accuracy of the network is minimally impacted. 
In this work we find a projection ${\bf P_{i}}$ that maps the features to a lower dimensional space by minimizing both the reconstruction error and the classification loss, as described in the rest of this section. 

\subsection{Single Layer Projection Compression}
\label{sub:rprCaP}
We first present how projection based compression is used to compress a single layer of a network.
To compress layer $i$, the output features are projected to low dimensional representation of rank $r$ using an orthonormal projection matrix ${\bf P_{i}}$ represented as a 4-Tensor of dimension $(c_{i+1} \! \times  \! r  \! \times  \! 1  \! \times  \!1 )$.
The optimal projection, ${\bf P^{*}_{i}}$ for layer $i$, based on minimizing the reconstruction loss is given as:
\begin{equation}
{\bf P^{*}_{i}} =\argmin_{{\bf P_{i}}} \norm{ {\bf O_{i}} \! - \! ({\bf I_{i}} \! * \! {\bf W_{i}} \! * \! {\bf P_{i}}+{\bf b_{i}} \! * \! {\bf P_{i}}) \! * \! {\bf P^{T}_{i}}}^{2}_{F}
\end{equation}
\noindent where $\norm{\cdot}^{2}_{F}$ is the Frobenious norm.

Inspired by \cite{zhang2015efficient}, we alter our optimization criteria to minimize the reconstruction loss of the input to the next layer.
This results in the optimization:
\begin{equation}
{\bf P^{*}_{i}} \! = \! \argmin_{{\bf P_{i}}} \norm{G({\bf O_i})\! - \! G(({\bf I_{i}} \! * \! {\bf W_{i}} \! * \! {\bf P_{i}} \! + \! {\bf b_{i}} \! * \! {\bf P_{i}}) \! * \! {\bf P^{T}_{i}})}^{2}_{F}
\end{equation}

\noindent The inclusion of the nonlinearity makes this a more difficult optimization problem.
In \cite{zhang2015efficient} the problem is relaxed and solved using Generalized SVD \cite{gower2004procrustes,takane2007regularized,takane2006generalized}. Our Cascaded Projection method is based on the end-to-end approach described next.

\begin{figure}[t]
\begin{center}
\includegraphics[width=3.25in]{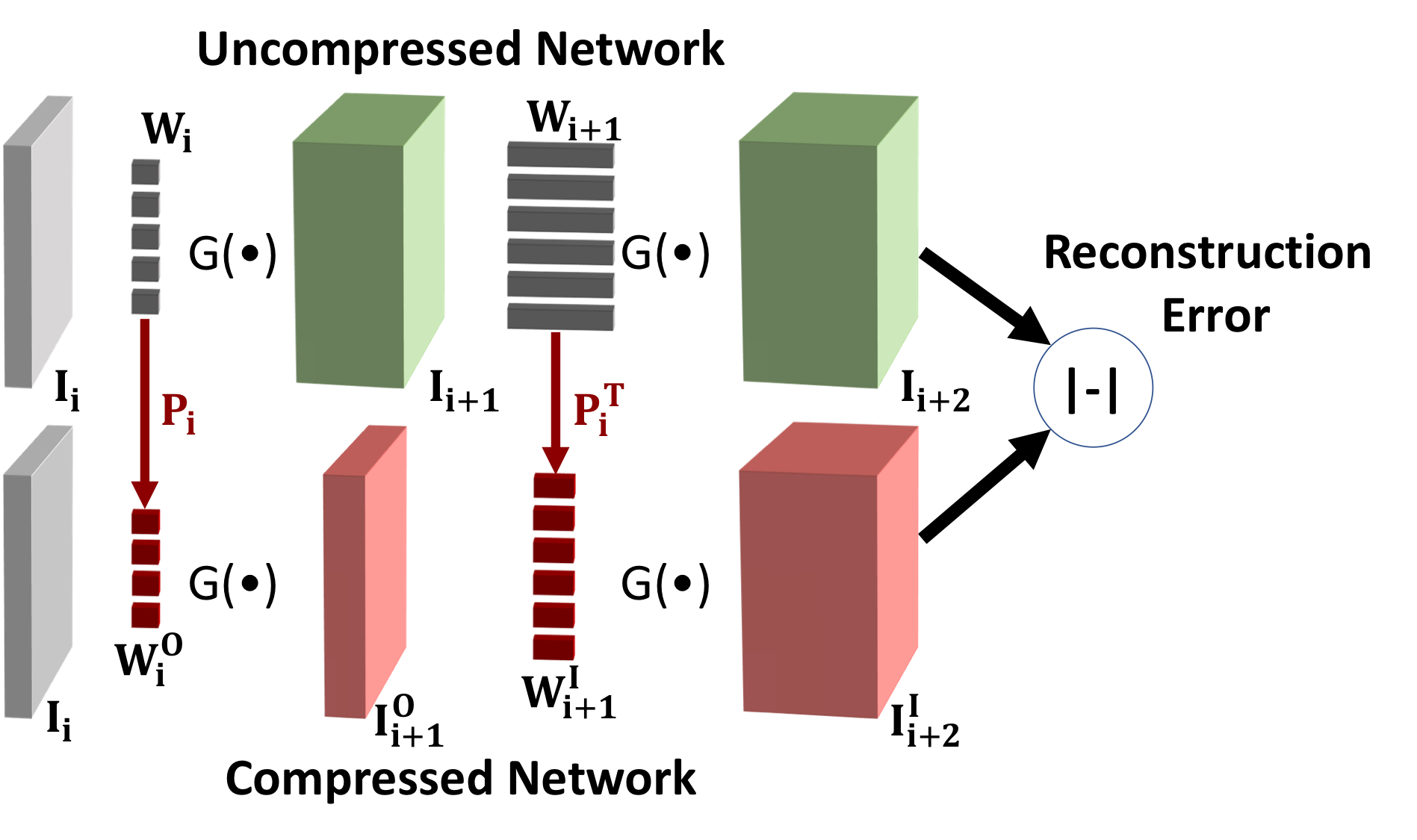}
\caption{Visual representation of the compression of a CNN layer using the CaP method to compress the filters 
${\bf W_{i}}$ and 
${\bf W_{i+1}}$ 
in the current and next layers 
using projections ${\bf P_{i}}$ and 
${\bf P_{i}^{T}}$
respectively. The reconstruction error in the next layer is computed after the nonlinearity $G(\cdot)$.
}
\label{fig:reconloss}
\end{center}
\end{figure}

\subsection{Cascaded Projection Compression}
\label{sub:cascadeCaP}
Factorization methods, including the single layer projection compression discussed above, are inefficient due to the additional convolution operations required to reproject the features to high dimensional space.  
Pruning methods
avoid reprojection 
by removing all channels associated with the pruned filters.
CaP takes a more powerful approach that forms linear combination of the kernels by projecting without the extra memory requirements of factorization methods.
Following the diagram in Figure \ref{fig:reconloss}, we consider two successive convolutional layers, labeled $i$ and $i \! + \! 1$, with kernels ${\bf W_{i}}$, ${\bf W_{i+1}}$ and biases ${\bf b_{i}}$, ${\bf b_{i+1}}$ respectively.
The input to layer $i$ is ${\bf I_{i}}$, while the output of layer $i \! + \! 1$ is the input to layer $i+2$, denoted by ${\bf I_{i+2}}$ and given below.
\begin{equation}
{\bf I_{i+2}}=G(G({\bf I_{i}}*{\bf W_{i}}+{\bf b_{i}})*{\bf W_{i+1}}+{\bf b_{i+1}})
\end{equation}
After substituting our compressed representation with reprojection for layer $i$ in the above we get:
\begin{equation}
{\bf I_{i+2}}=G(G(({\bf I_{i}}\!*\!{\bf W_{i}}\!*\!{\bf P_{i}}\!+\!{\bf b_{i}}\!*\!{\bf P_{i}})\!*\!{\bf P_{i}^{T}})\!*\!{\bf W_{i+1}}\!+\!{\bf b_{i+1}})
\end{equation}

To avoid reprojecting the low dimensional features back to higher dimensional space with ${\bf P_{i}^{T}}$, we seek two projections.
The first ${\bf P_{i}^{O}}$ which captures the optimal lower dimensional representation of the features in the current layer, and the second ${\bf P_{i}^{I}}$ which pulls the kernels of the next layer down to lower dimensional space. 
This formulation leads to an optimization problem over the projection operators:
\begin{equation}
\begin{aligned}
\{{\bf P^{I}_{i}}^{*},{\bf {P^{O}_{i}}^{*}}\}\! =\!
\argmin_{{\bf P^{I}_{i}},{\bf P^{O}_{i}}}  \lVert {\bf I_{i+2}}\!-\!G(G(({\bf I_{i}}\!*\!{\bf W_{i}}\!*\!{\bf P^{O}_{i}} \\
\!+\!{\bf b_{i}}\!*\!{\bf P^{O}_{i}}))\!*\!{\bf P^{I}_{i}}\!*\!{\bf W_{i+1}}\!+\!{\bf b_{i+1}})\rVert^{2}_{F}
\end{aligned}
\end{equation}

To make the problem tractable, we enforce two strong constraints on the projections. 
We require that they are orthonormal and transposes of each other:  ${\bf P_{i}^{I}}=({\bf {P^{O}_{i}}})^{T}$.
For the remainder of this work we replace ${\bf P_{i}^{O}}$ and ${\bf P_{i}^{I}}$ with ${\bf P_{i}}$ and ${\bf P_{i}^{T}}$, respectively.
These constraints make the optimization problem more feasible by reducing the parameter search space to a single projection operator for each layer. 
\begin{equation}
\begin{aligned}
{\bf P_{i}^{*}}\!=\!
\argmin_{{\bf P_{i}}, {\bf P_{i}}\in {\bf \mathbb{O}^{n\times m}}}  \lVert {\bf I_{i+2}}\!-\!G(G({\bf (I_{i}}\!*\!{\bf W_{i}}\!*\!{\bf P_{i}} \\
+{\bf b_{i}}\!*\!{\bf P_{i}}))\!*\!{\bf P^{T}_{i}}\!*\!{\bf W_{i+1}}\!+\!{\bf b_{i+1}})\rVert^{2}_{F}
\end{aligned}
\end{equation}

We solve the optimization of a single projection operator for each layer using a novel data-driven optimization method for projection operators discussed in Section \ref{sub:bptensor}.

\subsubsection{Kernel Compression and Relaxation }
\label{sub:kernelcomp}

\hspace{\parindent} Once the projection optimization is complete, we replace the kernels and biases in the current layer with their projected versions ${\bf W^{O}_{i}} \! = \! {\bf W_{i}} \! * \! {\bf P_{i}}$ and ${\bf b_{i}^{O}} \! = \! {\bf b_{i}} \! * \! {\bf P_{i}}$ respectively. 
We also replace the kernels in the next layer with their input compressed versions ${\bf W^{I}_{i+1}} \! = \! {\bf P_{i}^{T}} \! * \! {\bf W_{i+1}}$.  
Thus,
\begin{equation}
{\bf I_{i+2}}=G(G(({\bf I_{i}}*{\bf W^{O}_{i}}+{\bf b^{O}_{i}}))*{\bf W_{i+1}^{I}}+{\bf b_{i+1}})
\end{equation}

Figure \ref{fig:reconloss} depicts how the filters ${\bf W^{I}_{i+1}}$ in the next layer are compressed using the projection ${\bf P^{T}_{i}}$ and are therefore smaller than the kernels in the original network.
Utilizing the compressed kernels ${\bf W^{O}_{i}}$ and ${\bf W^{I}_{i}}$ results in twice the speedup over traditional factorization methods for all compressed intermediate layers (other than first and last layers).


Following kernel projection, we perform an additional round of training in which only the
compressed kernels are optimized. 
We refer to this step as kernel relaxation because we are allowing the kernels to find a better optimal solution after our projection optimization step. 

\subsection{Mixture Loss }
\label{sub:mixedloss}
A benefit of gradient based optimization is that a loss function can be altered to minimize both reconstruction and classification error.
Previous methods have focused on either reconstruction error minimization \cite{He_2017_ICCV,luo2017thin}  or classification \cite{Yu_2018_CVPR} based metrics when pruning each layer.
We propose using a combination of the standard cross entropy classification loss, $L_{Class}$, and the reconstruction loss $L_{R}$, shown in Figure \ref{fig:reconloss}.
The reconstruction loss for layer $i$ is given as:
\begin{equation}
\begin{aligned}
L_{R}(i) =  \lVert {\bf I_{i+2}}-G(G(({\bf I_{i}}*{\bf W_{i}}*{\bf P_{i}} \\
+{\bf b_{i}}*{\bf P_{i}}))*{\bf P^{T}_{i}}*{\bf W_{i+1}}+{\bf b_{i+1}})\rVert^{2}_{F}
\end{aligned}
\end{equation}
The mixture loss used to optimize the projections in layer $i$ is given as 
\begin{equation}
\label{eq:10}
\begin{aligned}
L(i) =  L_{R}(i) + \gamma L_{Class}
\end{aligned}
\end{equation}
\noindent where $\gamma$ is a mixture parameter that allows adjusting the impact of each loss during training. 
By using a combination of the two losses we obtain a compressed network that maintains classification accuracy while having feature representations for each layer which contain the maximal amount of information from the original network. 

\begin{figure*}[t]
\begin{center}
\includegraphics[width=0.95\textwidth]{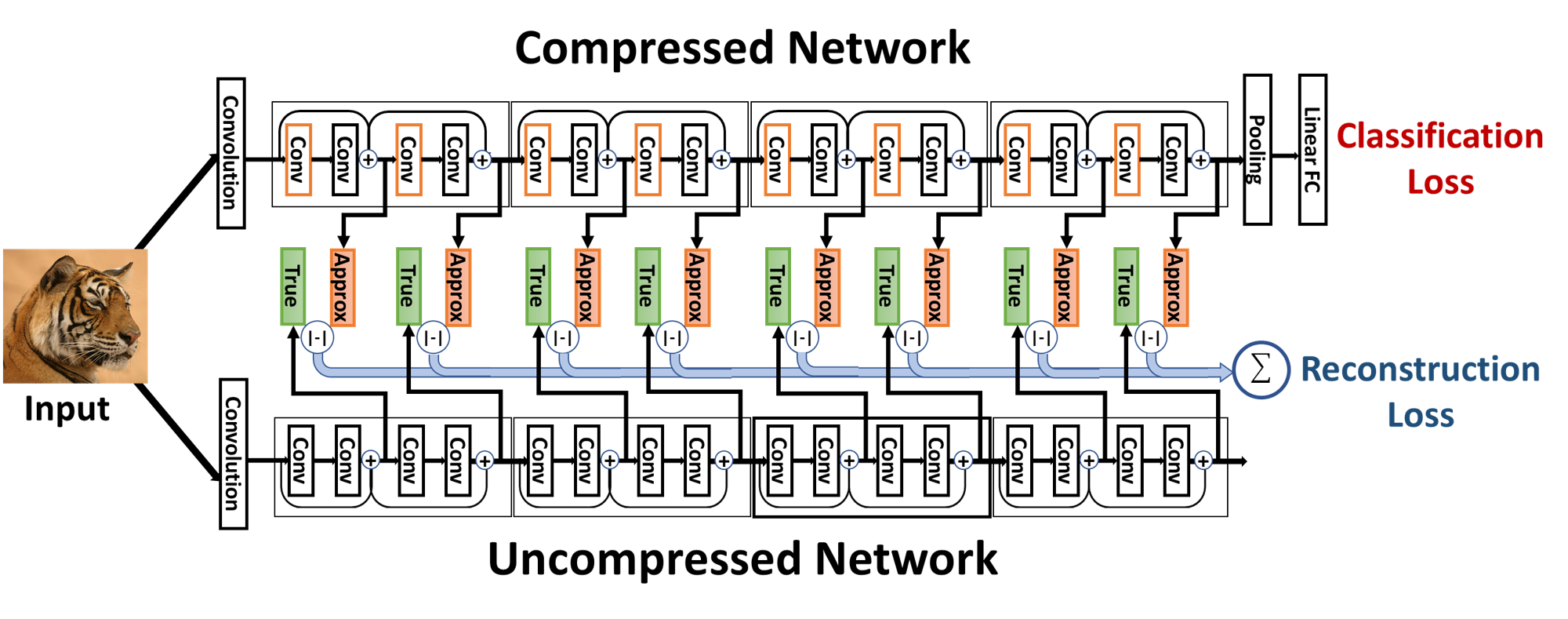}
\caption{Illustration of simultaneous optimization of the projections for each layer of the  ResNet18 network using a mixture loss that includes the classification loss and the reconstruction losses in each layer for intermediate supervision.
We do not alter the structure of the residual block outputs, therefore we do not affect residual connections and we do not compress the outputs of the last convolution layers in each residual block. }
\label{fig:fullComp}
\end{center}
\end{figure*}

\subsection{Compressing Multi-Branch Networks}
\label{sub:ResNet}

Multi-branch networks are popular due to their excellent performance and come in a variety of forms such as the Inception networks \cite{szegedy2015going, Szegedy16rethinking, szegedy2017inception}, Residual networks (ResNets) \cite{he2016deep} and Dense Networks (DenseNets) \cite{huang2017densely} among others. 
We primarily focus on applying CaP network compression to ResNets, but our method can be integrated with other multi-branch networks.
We select the ResNet architecture for two reasons. First, ResNets have a proven record of producing state-of-the art results \cite{he2016deep, he2016identity}.
And second, the skip connections work well with network compression, as they allow propagating information through the network regardless of the compression process within the individual layers. 

Our CaP modification for ResNet compression is illustrated in Figure \ref{sec:method}.
In our approach, we do not alter the structure of the residual block outputs, therefore we do not compress the outputs of the last convolution layers in each residual block, as was done by \cite{luo2017thin}.
In \cite{li2016pruning, Yu_2018_CVPR} pruning is performed on the residual connections, but we do not affect them, because pruning these layers has a large 
negative impact on the network's accuracy.

We calculate the reconstruction error in ResNets at the outputs of each residual block, as shown in Fig. \ref{fig:fullComp}, in contrast to single branch networks where we calculate the reconstruction error at the next layer as shown in Fig. \ref{fig:reconloss}.
By calculating the reconstruction error after the skip connections, we leverage the information in the skip connections in our projection optimization.

\subsubsection{Simultaneous Layer Compression}
\label{sub:layerfactor}
\hspace{\parindent}Most network compression methods apply a greedy layer-wise compression scheme, where one layer is compressed or pruned at a time. 
However, this layer-by-layer approach to network compression can lead to sub-optimal results \cite{Yu_2018_CVPR}.
We now present a version of CaP where all layers are simultaneously optimized.
This approach allows the latter layers to help guide the projections of the earlier layers and minimize the total reconstruction error throughout the network.

In our experiments, we found that simultaneous optimization of the projection matrices has the risk of becoming unstable when we compress more than one layer in each residual block. 
To overcome this problem we split the training of the projections in residual blocks with more than one compressible layer into two rounds.
In the first round, the projections for the odd layers are optimized, and in the second round the even layer projections are optimized. 

Additionally, we found that using the reconstruction loss at the final layers did not provide enough supervision to the network.
We therefore introduced deep supervision for each layer by minimizing the sum of normalized reconstruction losses for each layer, given by:
\begin{equation}
\label{eq:11}
\begin{aligned}
\argmin_{{\bf P_{i}} \in {\bf{P}}}\sum_{i=0}^{N}{L_{R}(i)}+\!\gamma L_{Class}
\end{aligned}
\end{equation}
\noindent where $\bf{P_i}$ is the projection for the $i^{th}$ layer, and $N$ is the total number of layers. 
We outline our approach to finding a solution for the above optimization using iterative backpropagation next.

\subsection{Back-Propagated Projection Optimization}
\label{sub:bptensor}


In this section we present an end-to-end Proxy Matrix Projection (PMaP) optimization method, which is an iterative optimization of the projection using  backpropagation with Stochastic Gradient Descent (SGD). 
The proposed method efficiently optimizes the network compression by combining  backpropagation with  geometric constraints.

In our framework, we restrict the projection operators to be orthogonal and thus satisfy ${\bf P_{i}}^{T}{\bf P_{i}}={\bf I}$.
The set of ($n \! \times \! m$) real-valued orthogonal matrices ${\bf \mathbb{O}^{n\times m}}$, forms a smooth manifold known as a Grassmann manifold. 
There are several optimization methods on Grassmann manifolds, most of which include iterative optimization and retraction methods  \cite{cunningham2015linear,  absil2009optimization, tagare2011notes,absil2012projection, wen2013feasible}. 

With CaP compression, the projection for each layer is dependent on the projections in all previous layers adding dependencies in the optimization across layers. 
Little work had been done in the field of optimization over multiple dependent Grassmann manifolds.
Huang \etal \cite{huang2017orthogonal} impose orthogonality constraints on the weights of a neural network during training using a method for  backpropagation of gradients through structured linear algebra layers developed in \cite{ivs_iccv15,ionescu2015training}.
Inspired by these works, we utilize a similar approach where instead of optimizing each projection matrix directly, we use a proxy matrix ${\bf X_i}$ for each layer $i$ and a transformation $\Phi(\cdot)$ such that $\Phi({\bf X_i})={\bf P_i}$.

We obtain the transformation $\Phi(\cdot)$ that projects each proxy matrix ${\bf X_{i}}$ to the closest location on the Grassmann manifold by performing Singular Value Decomposition (SVD) on ${\bf X_{i}}$, such that ${\bf X_{i}} \! = \! {\bf U_{i}}{\bf \Sigma_{i}}{\bf V_{i}^{T}}$, where 
${\bf U_{i}}$ and ${\bf V^{T}_{i}}$ are orthogonal matrices and ${\bf \Sigma_{i}}$ is the matrix of singular values.
The projection to the closest location on the Grassmann manifold is performed as $\Phi({\bf X_{i}}) \! = \! {\bf U_{i}}{\bf V^{T}_{i}}={\bf P_{i}}$.

During training, the projection matrix ${\bf P_{i}}$ is not updated directly; instead the proxy parameter ${\bf X_{i}}$ is updated based on the partial derivatives of the loss with respect to ${\bf U_{i}}$ and ${\bf V_{i}}$,
$\frac{\partial L }{\partial {\bf U_{i}}}$ and $\frac{\partial L}{\partial {\bf V_{i}}}$ respectively. 
The partial derivative of the loss $L$ with respect to the proxy parameter ${\bf X_{i}}$ was derived in \cite{ivs_iccv15,ionescu2015training} using the chain rule and is given by:
\begin{equation}
\label{eq:12}
\begin{aligned}
\frac{\partial L }{\partial {\bf X_{i}}} = {\bf U_{i}}\! \left\{ \! 2 {\bf \Sigma_{i}} \left( \!{\bf K^{T}_{i}} \! \circ \! \left( {\bf V^T_{i}} \frac{\partial L}{\partial {\bf V_{i}}}\right)\right)_{sym} \!+\!\frac{\partial L}{\partial {\bf \Sigma_{i}}}\right\}\!{\bf V^T_{i}}
\end{aligned}
\end{equation}

\noindent where $\circ$ is the Hadamard product, ${\bf A_{sym}}$ is the symmetric part of matrix A given as ${\bf A_{sym}} \! = \! \frac{1}{2}({\bf A^{T}} \! + \! {\bf A})$. Since $\Phi({\bf X_{i}}) \! = \! {\bf U_{i}}{\bf V^{T}_{i}}$, the loss does not depend on the matrix ${\bf \Sigma_{i}}$. Thus, $\frac{\partial L}{\partial {\bf \Sigma_{i}}} \! = \! 0$, 
and equation (\ref{eq:12}) becomes:
\begin{equation}
\begin{aligned}
\frac{\partial L }{\partial {\bf X_{i}}} = {\bf U_{i}}\! \left\{ \! 2 {\bf \Sigma_{i}} \left( \!{\bf K^{T}_{i}} \! \circ \! \left( {\bf V^T_{i}} \frac{\partial L}{\partial {\bf V_{i}}}\right)\right)_{sym}\right\}\!{\bf V^T_{i}}
\end{aligned}
\end{equation}

The above allows us to optimize our compression projection operators for each layer of the network using backpropagation and SGD.
Our method allows for end-to-end network compression using standard deep learning frameworks for the first time.

\section{Experiments}
\label{sec:exp}
We first perform experiments on independent layer compression of the VGG16 network to investigate how each layer responds to various levels of compression.
We then  perform a set of ablation studies on the proposed CaP algorithm to determine the impact for each step of the algorithm on the final accuracy of the compressed network. 
We compare CaP to other state-of-the-art methods by compressing the 
VGG16 network to have over 4$\times$ fewer floating point operations. 
Finally we present our experiments with varying levels of compression of ResNet architectures, with 18 or 50 layers, trained on the CIFAR10 dataset. 

All experiments were performed using PyTorch 0.4 \cite{paszke2017automatic} on a work station running Ubuntu 16.04. 
The workstation had an Intel i5-6500 3.20GHz CPU with 15 GB of RAM and a NVIDIA Titan V GPU.

\subsection{Layer-wise Experiments}
\label{sub:layerexp}

\begin{figure}[h]
\begin{center}
\includegraphics[width=2.4in]{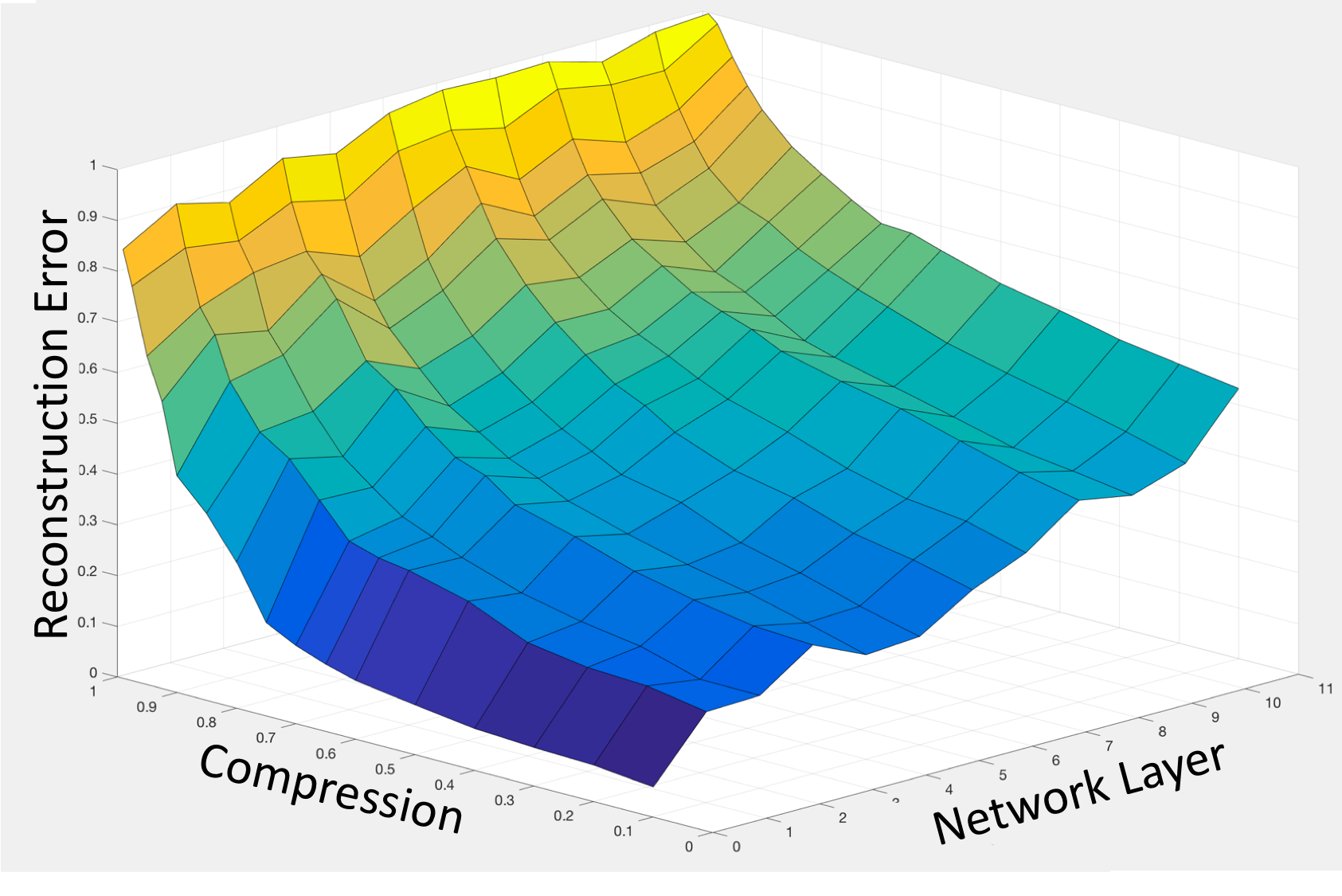}
\end{center}
\caption{Plot of the reconstruction error (vertical axis) for the range of compression (left axis) for each layer of the network (right axis). The reconstruction error is lower when early layers are compressed. }
\label{fig:reconPlot}
\end{figure}

\begin{figure}[h]
\begin{center}
\includegraphics[width=2.4in]{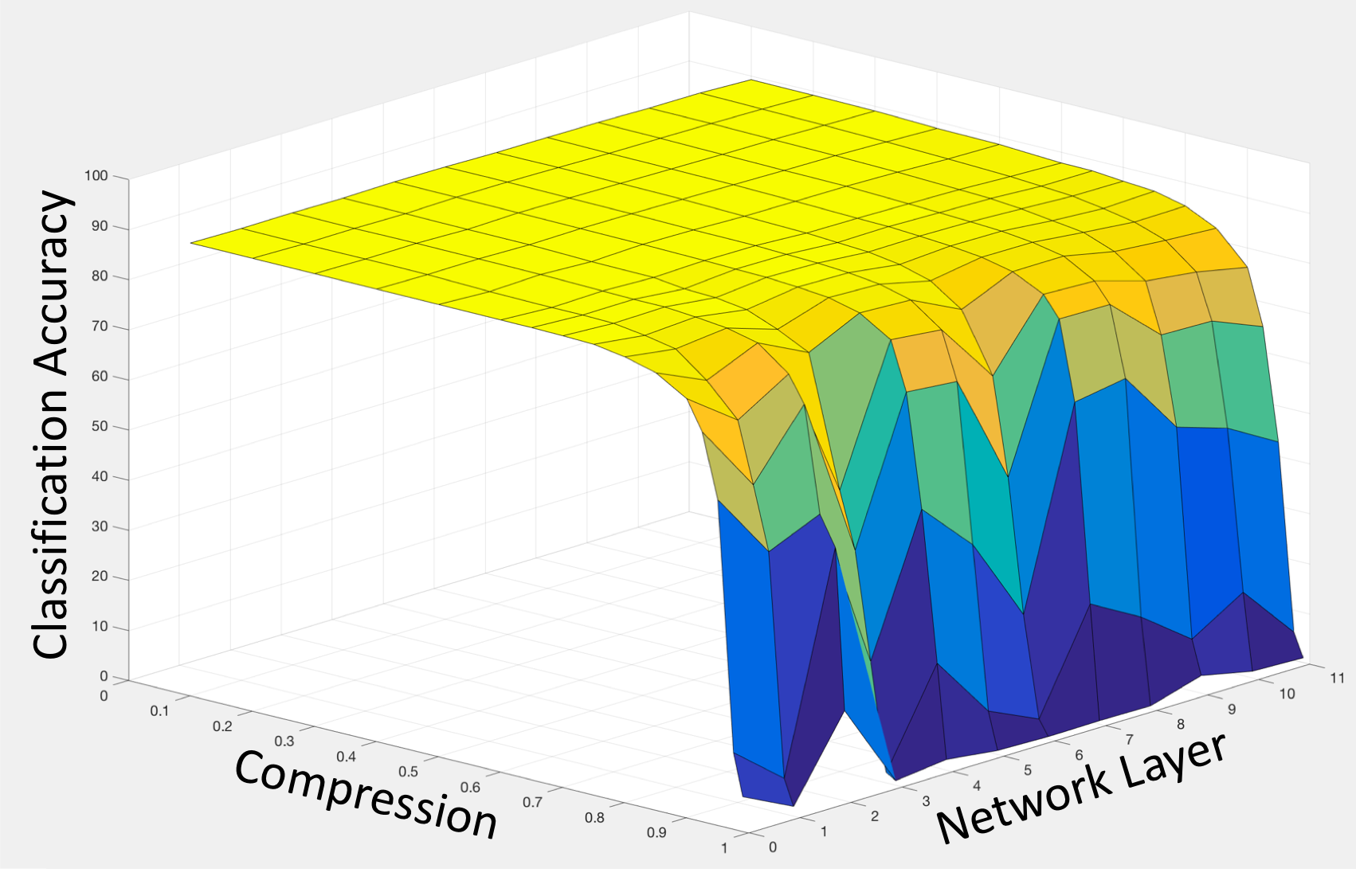}
\end{center}
\caption{Plot of the classification accuracy (vertical axis) for the range of compression (left axis) for each layer of the network (right axis).
The classification accuracy remains unaffected for large amounts of compression in a single layer anywhere in the network.
}
\label{fig:accPlot}
\end{figure}

In these experiment we investigate how each layer of the network is affected by increasing amounts of compression. 
We perform filter compression using CaP for each layer independently, while leaving all other layers uncompressed. 
We considered a range of compression for each layer, from 5\% to 99\%, and display the results in Figure 4.   
This plot shows two trends. 
Firstly the reconstruction error does not increase much until 70\% compression, indicating that a large portion of the parameters in each layer are redundant and could be reduced without much loss in accuracy.
The second trend is the increase in reconstruction error for each level of compression for the deeper layers of the network (right axis). 

In Figure \ref{fig:accPlot} we plot the network accuracy resulting from each level of compression for each layer. 
The network is relatively unaffected for a large range of compression, despite the fact that there is a significant amount of reconstruction error introduced by the compression shown in Figure \ref{fig:reconPlot}.

\subsection{CaP Ablation Experiments}
\label{sub:compContExp}
We ran additional experiments to determine the contribution of the projection optimization and kernel relaxation steps of our algorithm.
We first trained the ResNet18 network on the CIFAR100 dataset and achieved a baseline accuracy of 78.23\%.
We then compressed the network to 50\% of the original size using only parts of the CaP method to assess the effects of different components.
We present these results in Table \ref{tbl:res18cifar100}.

We also trained a compressed version ResNet18 from scratch for 350 epochs, to provide a baseline for the compressed ResNet18 network.
When only projection compression is performed on the original ResNet18 network, there was a drop in accuracy of 1.58\%.
This loss in classification accuracy decreased to  0.76\% after kernel relaxation. 
In contrast, when the optimized projections are replaced with random projections and only kernel relaxation training is performed, there is a 1.96\% drop in accuracy, a 2.5 times increase in classification error.
These results demonstrate that the projection optimization is an important aspect of our network compression algorithm, and the combination of both steps outperforms training the compressed network from scratch. 

\begin{table} [!h]
\begin{center}
\begin{tabular}{|c|c|c|c|}
\hline
\!{\bf ResNet18 Network Variation}\! & {\!\bf Accuracy\!}\\ 
\hline
ResNet18 Uncompressed (upper bound) & 78.23\\
\hline
Compressed ResNet18 from Scratch & 77.22\\
\hline
CaP Compression with Projection Only & 76.65\\
\hline
CaP with Random Proj. \& Kernel Relax & 76.27\\
\hline
CaP with Projection \& Kernel Relax & {\bf77.47}\\
\hline
\end{tabular}
\end{center}
\caption{Network compression ablation study of the CaP method compressing the ResNet18 Network trained on the CIFAR100 dataset. ({Bold numbers} are best).}
\label{tbl:res18cifar100}
\end{table} 

\begin{figure}[h]
\begin{center}
\includegraphics[width=1.8in]{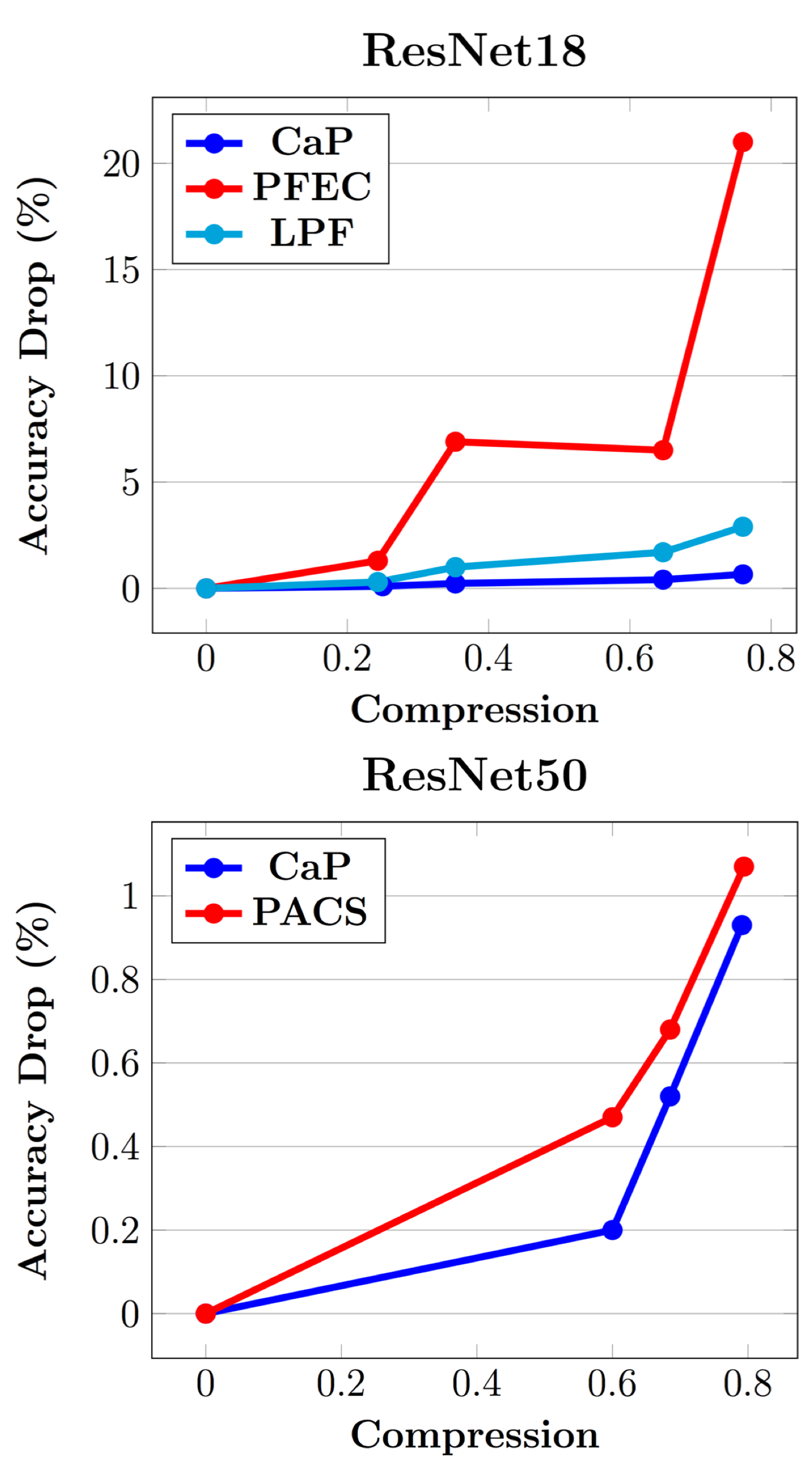}
\end{center}
\caption{Classification accuracy drop on CIFAR10, relative to baseline, of compression methods (CaP, PCAS \cite{yamamoto2018pcas}, PFEC \cite{li2016pruning} and LPF \cite{huang2018learning})  for a range of compression levels on ResNet18 (Top) and ResNet50 (Bottom).  }
\label{fig:ResAccPlot}
\end{figure}


\begin{table} [t]
\begin{center}
\begin{tabular}{|c|c|c|c|c|c|}
\hline
\!{\bf ResNet}\! & {\bf Method}  &{\bf FT}  & {\bf\!FLOPs \!\%} & {\bf Acc. / Base} \\
\hline
\multirow{11}{*}{56} &PFEC \cite{li2016pruning}& N & 72.4 & 91.31 / 93.04\\
&CP \cite{He_2017_ICCV}& N & 50.0 & 90.90 / 92.80\\
&SFP \cite{he2018soft}  & N & 47.4 & 92.26 / 93.59\\
&AMC \cite{he2018amc} & N & 50.0 & 90.1 / 92.8\\
&CaP & N & 50.2 &   {\bf92.92} / 93.51\\
\cline{2-5}
&PFEC \cite{li2016pruning}&Y& 72.4 & 93.06 / 93.04\\
&NISP \cite{Yu_2018_CVPR} &Y& 57.4 & (-0.03) *\\
&CP \cite{He_2017_ICCV}&Y& 50.0 & 91.80 / 92.80\\
&SFP \cite{he2018soft} &Y & 47.4 & {\bf93.35} / 93.59\\
&AMC \cite{he2018amc}&Y & 50.0 & 91.9 / 92.8\\
&CaP&Y & 50.2 &   93.22 / 93.51\\
\hline
\hline
\multirow{8}{*}{110}&PFEC \cite{li2016pruning}& N &61.4 & 92.94 / 93.53\\
&MIL \cite{dong2017more} &N & 65.8 & 93.44 / 93.63\\
&SFP \cite{he2018soft} &N & 59.2 & 93.38 / 93.68\\
&CaP& N& 50.1  &  {\bf 93.95}/ 94.29\\
\cline{2-5}
&PFEC \cite{li2016pruning}& Y&61.4 & 93.30 / 93.53\\
&NISP \cite{Yu_2018_CVPR} &Y& 56.3 & (-0.18) *\\
&SFP \cite{he2018soft}  &Y& 59.2 & 93.86 / 93.68\\
&CaP &Y& 50.1 &  {\bf 94.14}/ 94.29\\
\hline
\end{tabular}
\end{center}
\caption{Comparison of CaP with pruning and factorization based methods using ResNet56 and ResNet110 trained on CIFAR10. FT denotes fine-tuning. ({Bold numbers} are best). * Only the relative drop in accuracy was reported in \cite{Yu_2018_CVPR} without baseline accuracy.} 
\label{tbl:res56-110cifar10}
\end{table}

\subsection{ResNet Compression on CIFAR 10}
\label{sub:resnetcifar}

We perform two sets of experiments using ResNet18 and ResNet50 trained on the CIFAR10 dataset \cite{krizhevsky2009learning}.
We compress 18 and 50 layer ResNets with varying levels of compression and compare the relative drop in accuracy of CaP with other state-of-the-art methods \cite{yamamoto2018pcas,li2016pruning,huang2018learning}.
We plot the drop in classification accuracy for ResNet18 and ResNet50 in Figure \ref{fig:ResAccPlot}.
For both networks, the CaP method outperforms the other methods for the full range of compression.

In Table \ref{tbl:res56-110cifar10}, we present classification accuracy of ResNet56 and ResNet110 with each residual block compressed to have 50\% fewer FLOPs using CaPs. 
We compare the results obtained by CaP with those of  \cite{he2018soft, he2018amc, li2016pruning, Yu_2018_CVPR, He_2017_ICCV} where the networks have been subjected to similar compression ratios. 
We report accuracy results with and without fine-tuning and include the baseline performance for comparison.  

Results with fine-tuning are generally better, except in cases when there is over-fitting.  However, fine-tuning for a long period of time can hide the poor performance of a compression algorithm by retraining the network filters away from the compression results.
The results of the CaP method without fine-tuning are based on projection optimization and kernel relaxation on the compressed filters with reconstruction loss, while the fine-tuning results are produced with an additional round of training based on mixture loss for all of the layers in the network.

\subsection{VGG16 Compression with ImageNet}
\label{sub:IamgeNetVGG}

\begin{table*} [!h]
\begin{center}
\begin{tabular}{|c|c|c|c|c|c|}
\hline
{\bf Method} & {\bf Parameters} & {\bf Memory (Mb)} & {\bf FLOPs} & {\bf GPU Speedup} & {\bf Top-5 Accuracy / Baseline} \\
\hline
VGG16 \cite{sz15} (Baseline) & 14.71M & 3.39 & 30.9B & 1 & 89.9\\
\hline
Low-Rank \cite{jaderberg2014speeding} & - & - & - & 1.01* & 80.02 / 89.9\\
\hline
Asym. \cite{zhang2015efficient} & \bf{5.11M} & 3.90  & 3.7B  & 1.55* & 86.06  / 89.9\\
\hline
Channel Pruning \cite{He_2017_ICCV} & 7.48M  & 1.35 & \bf{6.8B} & 2.5* & 82.0 / 89.9\\
\hline
CaP (based on \cite{He_2017_ICCV} arch) & 7.48M  & 1.35  &\bf{6.8B } & 3.05 & 86.57 / 90.38\\
\hline
CaP Optimal & 7.93M  & \bf{1.11 } & \bf{6.8B } &  \bf{3.44} & \textbf{88.23} / 90.38\\
\hline
\end{tabular}
\end{center}
\caption{Network compression results of pruning and factorization based methods without fine-tuning. 
The top-5 accuracy of the baseline VGG16 network varies slightly for each of the methods due to different models and frameworks.
({Bold numbers are best}). Results marked with * were obtained from \cite{He_2017_ICCV}.} 
\label{tbl:VGG16noFT}
\end{table*}

We compress the VGG16 network trained on ImageNet2012 \cite{deng2009imagenet} and compare the results of CaP with other state-of-the-art methods. 
We present two sets of results, without fine-tuning and with fine-tuning, in Tables \ref{tbl:VGG16noFT} and \ref{tbl:VGG16FT} respectively.
Fine-tuning on  ImageNet  is time intensive  and requires significant computation power.  This is a hindrance for many applications where users do not have enough resources to retrain a compressed network. 

\begin{table} [!t]
\begin{center}
\begin{tabular}{|c|c|c|c|}
\hline
\multirow{2}{*}{\bf Method} & {\bf Mem.} & \multirow{2}{*}{\bf FLOPs} & {\bf Top-5 Acc. } \\ & ({\bf Mb}) &&{ \bf / Baseline}\\
\hline
VGG16 \cite{sz15} & 3.39 & 30.9B & 89.9\\
\hline
Scratch \cite{He_2017_ICCV} & 1.35 &  6.8B & 88.1 \\
\hline
COBLA \cite{li2018constrained} & 4.21 & 7.7B & 88.9 / 89.9\\
\hline
Tucker \cite{kim2015compression} & 4.96 & \textbf{6.3B} & \textbf{89.4} / 89.9\\
\hline
CP \cite{He_2017_ICCV} &  1.35 & 6.8B & 88.9 / 89.9 \\
\hline
ThiNet-2 \cite{luo2017thin} &  {1.44} &  6.7B & 88.86 / 90.01 \\
\hline
CaP  & \bf{1.11} & 6.8B & 89.39 / 90.38\\
\hline
\end{tabular}
\end{center}
\caption{Network compression results of pruning and factorization based methods with fine-tuning. ({Bold numbers} are best). }
\label{tbl:VGG16FT}
\end{table} 

In Table \ref{tbl:VGG16noFT} we compare CaP with factorization and pruning methods, all without fine-tuning.  
As expected, factorization methods suffer from increased memory load due to their additional intermediate feature maps. 
The channel pruning method in \cite{He_2017_ICCV} has a significant reduction in memory consumption but under-performs the factorization method in \cite{zhang2015efficient} without fine-tuning. 
We present two sets of results for the CaP algorithm, each with different levels of compression for each layer. 
To match the architecture used in \cite{He_2017_ICCV} we compressed layers 1-7 to 33\% of their original size, and filters in layers 8-10 to 50\% of their original size, while the remaining layers are left uncompressed .
We also used the CaP method with a compression architecture that was selected based on our layer-wise training experiments.
The results in Table \ref{tbl:VGG16noFT} demonstrate that the proposed CaP compression achieves higher speedup and higher classification accuracy than the factorization or pruning methods. 

In Table \ref{tbl:VGG16FT} we compare CaP with state-of-the-art network compression methods, all with fine-tuning. 
The uncompressed  VGG16 results are from \cite{sz15}.
We include results from training a compressed version of VGG16 from scratch on the ImageNet dataset as reported in \cite{He_2017_ICCV}. 
We compare CaP with the results of two factorization methods \cite{li2018constrained, kim2015compression} and two pruning methods \cite{He_2017_ICCV}, \cite{luo2017thin}.
Both factorization methods achieve impressive classification accuracy, but this comes at the cost of increased memory consumption. 
The pruning methods reduce both the FLOPs and the memory consumption of the network, while maintaining high classification accuracy.
However, they rely heavily on fine-tuning to achieve high accuracy. 
We lastly provide the results of the CaP compression optimized at each layer. 
Our results demonstrate that the CaP algorithm gives state-of-the-art results, has the largest reduction in memory consumption, and outperforms the pruning methods in terms of top-5 accuracy. 

\section{Conclusion}
\label{sec:conc}

In this paper, we propose cascaded projection, an end-to-end trainable framework for network compression that optimizes compression in each layer.
Our CaP approach forms linear combinations of kernels in each layer of the network in a manner that both minimizes reconstruction error and maximizes classification accuracy.
The CaP method is the first in the field of network compression to optimize the low dimensional projections of the layers of the network using  backpropagation and SGD, using our proposed Proxy Matrix Projection optimization method.

We demonstrate state-of-the-art performance compared to pruning and factorization methods, when the CaP method is used to compress standard network architectures trained on standard datasets.
A side benefit of the CaP formulation is that it can be performed using standard deep learning frameworks and hardware, and it does not require any specialized libraries for hardware for acceleration. 
In future work, the CaP method can be combined with other methods, such as quantization and hashing, to further accelerate deep networks. 

\clearpage

{\small
\bibliographystyle{ieee}
\bibliography{CascadedProjection}
}

\end{document}